\documentclass[conference,letterpaper]{IEEEtran}
\IEEEoverridecommandlockouts

\usepackage[bottom=2.5cm,top=2.2cm,left=1.6cm,right=1.6cm]{geometry}
\usepackage{booktabs}
\usepackage{threeparttable}
\usepackage{amsmath,graphicx,cite,amssymb}
\usepackage{amsfonts,multirow,bm,array,setspace}

\usepackage{graphicx,float,cite,amssymb}
\usepackage{multirow,bm,array,setspace}
\usepackage{textcomp}
\usepackage{psfrag}
\usepackage{color}
\usepackage{pstricks}
\usepackage{bbm}
\usepackage{diagbox}
\usepackage{enumitem}
\usepackage{algorithm,algpseudocode}

\newcommand{\bx}{\bm x}

\def\BibTeX{{\rm B\kern-.05em{\sc i\kern-.025em b}\kern-.08em
    T\kern-.1667em\lower.7ex\hbox{E}\kern-.125emX}}
\begin{document}

\title{Crossword: A Semantic Approach to Data Compression via Masking\\
}

\author{
    \IEEEauthorblockN{Mingxiao Li$^{\dag}$, Rui Jin$^{\dag}$, Liyao Xiang$^{\S}$, Kaiming Shen$^\dag$, and Shuguang Cui$^\dag$}
    \IEEEauthorblockA{$^\dag$School of Science and Engineering, FNii, The Chinese University of Hong Kong (Shenzhen), China\\ 
    $^\S$Shanghai Jiao Tong University, China\\
    E-mail: \{mingxiaoli, ruijin\}@link.cuhk.edu.cn; xiangliyao08@sjtu.edu.cn;\\
    shenkaiming@cuhk.edu.cn; shuguangcui@cuhk.edu.cn
}}

\maketitle

\begin{abstract}

The traditional methods for data compression are typically based on the symbol-level statistics, with the information source modeled as a long sequence of i.i.d. random variables or a stochastic process, thus establishing the fundamental limit as entropy for lossless compression and as mutual information for lossy compression. However, the source (including text, music, and speech) in the real world is often statistically ill-defined because of its close connection to human perception, and thus the model-driven approach can be quite suboptimal. This study places careful emphasis on English text and exploits its semantic aspect to enhance the compression efficiency further. The main idea stems from the puzzle crossword, observing that the hidden words can still be precisely reconstructed so long as some ``key'' letters are provided. The proposed masking-based strategy resembles the above game. In a nutshell, the encoder evaluates the semantic importance of each word according to the semantic loss and then masks the minor ones, while the decoder aims to recover the masked words from the semantic context by means of the Transformer. Our experiments show that the proposed semantic approach can achieve much higher compression efficiency than the traditional methods such as Huffman code and UTF-8 code, while preserving the meaning in the target text to a great extent.
\end{abstract}


\section{Introduction}

Data compression, or source coding in a communication system, has been widely recognized as an artful trick that involves human perception, especially for sources like literature, music, speech etc. Indeed, Shannon asserted in his seminal paper \cite{shannon1948mathematical} that ``these semantic aspects of communication are irrelevant to the engineering problem'' assuming that
\begin{enumerate}[label=(\roman*)]
    \item the source can be modeled as a stochastic process or even a sequence of i.i.d. random variables;
    \item the source comprises infinitely many symbols.
\end{enumerate}
However, neither of the above assumptions may hold in practice. This work proposes a semantic approach to the English text compression, which improves upon the traditional model-based approach by taking the meaning of words into account. The main idea is to mask those semantically less important words throughout the text, thus enhancing the compression efficiency while preserving the overall meaning of the text.

The proposed masking strategy for semantic data compression is inspired by the popular puzzle \emph{crossword}---the goal of which is to fill the blank grids with letters and thus form words or phrases in accordance with the given grid layout and the fixed grids already filled with letters. It turns out that the above word puzzle has an analogy to data compression. The encoder can be thought of as the designer of the puzzle, who wishes to mask as many words as possible (i.e., maximize the portion of blank grids) to increase the difficulty of the puzzle. In the meanwhile, the decoder acts as the player and aims to recover all the words successfully. We seek the optimal word masking that attains an equilibrium satisfying both the puzzle designer and the player. Furthermore, as the main feature of this puzzle, words cross each other; it can be perceived that the letters lying in those intersection grids are often easier to guess. Likewise, with regard to the English text compression, some particular words can be identified as the ``intersection grid'' type if their removals do not cause much semantic loss, and thus can be masked without undermining the decoder's capability to recover the text. Notice that the choice of such words typically depends on the context, e.g., the word ``affairs'' is very likely to be semantically minor if the sample text is taken from a political document owned by the European Parliament, as observed in our experiments. The cutting-edge artificial intelligence tools such as the Bidirectional Encoder Representation (BERT) \cite{devlin2018bert,reimers2019sentence} and the Transformer \cite{vaswani2017attention} are the critical enabler for sensing the background knowledge, detecting the semantically minor words, and preserving the semantic information.

The efforts in semantic compression date back to \cite{witten1994semantic} in the 1990s, wherein a simple idea of replacing every word with a shorter equivalent from the thesaurus is discussed. The authors of \cite{witten1994semantic} further discuss how to mimic the genre (e.g., Hemingway's) in a generative fashion. Another early attempt can be found in \cite{isi2001semantic} with the attention focused on the Extensible Markup Language (XML). Its primary idea is to classify the XML data according to their importance, and then reduce the precision of those data that can tolerate higher loss. Moreover, \cite{basu2014preserving} suggests defining the semantic entropy to be mutual information between the word model and syntactic message, and then uses that to measure the limit of semantic compression. Another line of studies \cite{wang2022sinc,bao2011towards}
use the connection between facts from the knowledge basis and hence reduce the redundancy in a semantic sense. The tool of sentence-BERT \cite{reimers2019sentence} has been applied extensively in the past few years for the word embedding purpose in a variety of semantic communication tasks ranging from source coding \cite{vial2019sense,niu2022paradigm} to joint source-channel coding \cite{qin2021semantic,dai2022communication}. In contrast to \cite{vial2019sense} that aims at data compression on a per-word basis, our work adopts the sentence-BERT for data compression at a higher level that encompasses the overall context. In particular, with the revival of neural networks, the most recent trend in the area is towards deep learning-based methodology, such as \cite{qin2021semantic} that considers resource allocation, \cite{hu2022robust} that considers the robust transmission against semantic noise, and \cite{xie2021task} that considers multi-user semantic communications.


\section{System Model}

Consider an English text denoted by $T$ that comprises a total of $M$ sentences:
\begin{equation}
    T = (S_1,S_2,\ldots,S_M),
\end{equation}
where $S_i$, $i=1,\ldots,M$, is the $i$th sentence. Assume that these sentences may vary in length. For the data compression purpose, $T$ is converted into $\bx$ in the set of finite-length bit strings $\{0,1\}^*$ by the encoding function $f(\cdot)$ as
\begin{equation}
    \bx = f(T)\in\{0,1\}^*.
\end{equation}
Conversely, the decoding function $g(\cdot)$ aims to recover the text $T$ from the bit string $\bx$ as
\begin{equation}
   \widehat T = g(\bx).
\end{equation}
Moreover, denoting by $\widehat S_i$ the $i$th recovered sentence, we can split the decoded text into
\begin{equation}
    \widehat{T} = (\widehat S_1,\widehat S_2,\ldots,\widehat S_M).
\end{equation}
Let $\delta(T,\widehat T)\ge0$ be the distortion cost it incurs for representing the ground truth $T$ by the decoded text $\widehat T$. As a typical problem formulation of data compression, we seek the optimal encoder-and-decoder pair $(f,g)$ that minimizes the length of $\bx$, which is denoted by $\mathrm{len}(\bx)$, under the distortion constraint $\epsilon$, i.e.,
\begin{subequations}
  \label{prob:FP_min:quadratic}
\begin{align}
&\underset{f(\cdot),\;g(\cdot)}{\text{minimize}} \quad\; \mathrm{len}(\bx)
  \label{prob:FP_min:quadratic:obj}\\
&\text{subject to}\quad \delta(T,\widehat{T})\le \epsilon.
  \label{prob:FP_min:quadratic:cons_c}
\end{align}    
\end{subequations}
Notice that $T$ is not fixed \emph{a priori} in the above problem, so we aim at a universal design of $f(\cdot)$ and $g(\cdot)$ that works universally for all possible texts.

It remains to specify the distortion function $\delta(T,\widehat T)$. In classical studies based on the classical rate-distortion theory, $T$ and $\widehat T$ are typically compared on a per-symbol basis; the symbol in English text is either a letter or a punctuation mark. Given the $j$th symbol $\ell_j$ from $T$ and the $j$th symbol $\hat\ell_j$ from $\widehat T$, the traditional method computes the symbol-level distortion $d(\ell_{j},\hat\ell_j)$, a common example of which is Hamming distortion:
\begin{equation}
    d(\ell_{j},\hat\ell_j)=\begin{cases}
      0 & \text{if $\ell_j=\hat\ell_j$},\\
      1 & \text{otherwise}.
    \end{cases}
\end{equation}
Then the overall text distortion $\delta(T,\widehat T)$ amounts to the sum of the symbol distortions $d(\ell_{j},\hat\ell_j)$ for all $j$. In principle, the above strategy takes each symbol as the random variable unit and thus treats the text as a sequence of random variables. The resulting data compression algorithm can only take advantage of the language statistics at the symbol level.

Rather, this work proposes to evaluate the distortion on the inter-word and inter-sentence context basis, thereby capturing the semantic aspect of the target text. Consider a pair of sentences $S_i$ and $\widehat S_i$. First, convert them to two Euclidean vectors in $\mathbb R^{384}$ by means of the sentence-BERT \cite{reimers2019sentence}, i.e.,
\begin{equation}
\label{bert}
    \bm\mu_i = \texttt{BERT}(S_i)\quad\text{and}\quad \hat{\bm\mu}_i = \texttt{BERT}(\widehat S_i).
\end{equation}
Then, in order to quantify the goodness of representing $S_i$ by $\widehat S_i$, we use the \emph{cosine similarity} between their respective sentence-BERT vectors
\begin{equation}
\label{eta}
\eta(S_i,\widehat{S}_i)=
\frac{\bm\mu_i^\top\hat{\bm\mu}_i}{\|\bm\mu_i\|_2 \cdot\|\hat{\bm\mu}_i\|_2}
\end{equation}
and further consider the \emph{cosine distance}
\begin{equation}
\label{lambda}
\lambda(S_i,\widehat{S}_i)=
1-\eta(S_i,\widehat{S}_i).
\end{equation}
Thus, from a semantic perspective, we suggest the following definition of $\delta(T,\widehat T)$:
\begin{equation}
\label{delta}
    \delta(T,\widehat T) = \sum^M_{i=1}\alpha_i\lambda(S_i,\widehat{S}_i),
\end{equation}
where the positive weight $\alpha_i$ reflects the semantic importance of $S_i$ in terms of the overall text $T$. We set $\alpha_i=N_i$ out of the belief that the semantic information is proportional to the sentence length. 


\begin{figure}[t]

\centering

\includegraphics[width=8.9cm]{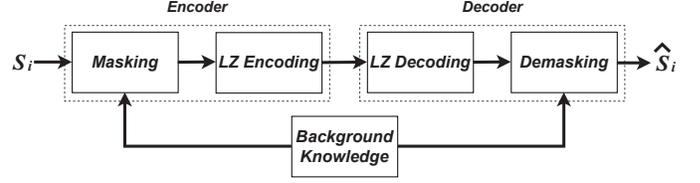}

\caption{Paradigm of the proposed masking-based semantic data compression.}
\label{fig:system}
\end{figure}


\begin{algorithm}[t]
\caption{Semantic Data Compression (Encoder Part)}
\label{alg:EN}
\begin{algorithmic}[1]
\Statex \textbf{Input:} English text $T$ and masking ratio $\rho\in[0,1)$
\State \textbf{Initialization:} Word list $\mathcal K=\emptyset$
\For{each sentence $S_i$ in $T$}
    \State Compute $\bm\mu_i=\texttt{BERT}(S_i)$
    \For{each word $W_{in}$ in $S_i$}
        \State Update $\mathcal K=\mathcal K\cup\{W_{in}\}$
        \State Compute $\hat{\bm\mu}_i=\texttt{BERT}(S^-_{in})$
        \State Compute $ \sigma_{in} =
        1-{\bm\mu_i^\top\hat{\bm\mu}_i}/({\|\bm\mu_i\|_2 \cdot\|\hat{\bm\mu}_i\|_2})$
    \EndFor
    \State Normalize each $\sigma_{in}$ as $\bar\sigma_{in}$ according to \eqref{norm_sigma}
\EndFor
\For{each word $V_k\in\mathcal K$}
    \State compute the weight $\mu_k$ according to \eqref{mu}
\EndFor
\State Sort the words in $\mathcal K$ in the ascending order of $\mu_k$
\State Replace the top $\lfloor\rho|\mathcal K|\rfloor$ words with $\textbf{\#}$ throughout $T$
\State Concert the masked text to the bit string $\bx$ by LZ encoding
\Statex \textbf{Output:} Bit string $\bx$
\end{algorithmic}
\end{algorithm} 

\section{Main Results}

The overall paradigm of the proposed semantic approach to data compression is visualized in Fig.~\ref{fig:system}. The encoding part and the decoding part are specified in the sequel, then followed by a discussion as to why the traditional entropy metric does not work for this paradigm.

\subsection{Semantic Encoding}

The proposed semantic encoder consists of two parts: masking followed by bit-string conversion. Assume that the sentence $S_i$ has $N_i$ words. With the $n$th word denoted by $W_{in}$, the sentence $S_i$ can be expressed as
\begin{equation}
    S_i = (W_{i1},W_{i2},\ldots,W_{iN_i}).
\end{equation}
For each $n=1,\ldots,N_i$, we replace the current $W_{in}$ with \# at a time in sentence $S_{i}$; the resulting new sentence is denoted by $S^-_{in}$. We evaluate the corresponding semantic loss as
\begin{equation}
    \sigma_{in} = \lambda(S_{i},S^-_{in}).
\end{equation}
After obtaining all the $\sigma_{in}$'s within the sentence $S_i$, we further normalize the semantic loss:
\begin{equation}
\label{norm_sigma}
    \bar\sigma_{in} = \frac{\sigma_{in}}{\sum^{N_i}_{n'=1}\sigma_{in'}}.
\end{equation}
Intuitively, $\bar\sigma_{in}$ indicates the portion of the total semantic information of $S_i$ taken on by $W_{in}$ alone. The above procedure is carried out for every sentence $S_i$.

Subsequently, we collect all the words that have appeared in the text together and sort them (e.g., in lexicographic order) in the list 
\begin{equation}
    \mathcal K = \left\{V_1, V_2\ldots,V_{|\mathcal K|}\right\},
\end{equation}
where the $V_k$'s are the distinct words from $T$. We use the set $\mathcal Q_{ik}$ to record the positions of all the occurrences of the word $V_k$ in the sentence $S_i$; notice that the same word may occur multiple times within one sentence. In particular, $\mathcal Q_{ik}=\emptyset$ if $S_i$ does not contain $V_k$. For each word $V_k$ in the list, its semantic importance in terms of the present text is quantified by the weight
\begin{equation}
\label{mu}
    \mu_k = \frac{1}{\sum^M_{i=1}|\mathcal Q_{ik}|}\cdot\sum^M_{i=1}\left(\alpha_i\sum_{n\in\mathcal Q_{ik}}\bar\sigma_{in}\right).
\end{equation}
Sort all the words in the list in the ascending order of $\mu_k$. For some given \emph{masking ratio} $0\le\rho<1$, replace the top $\lfloor \rho |\mathcal K|\rfloor$ words with the special symbol ``$\textbf{\#}$'' throughout the text $T$, namely \emph{masking}. As the final step in encoding, we convert the masked text to the bit string by the standard method, e.g., Lempel-Ziv (LZ) code \cite{lempel1976complexity}. Algorithm \ref{alg:EN} summarizes the above encoding procedure.

\begin{figure}[t]
\centering

\includegraphics[width=8.9cm]{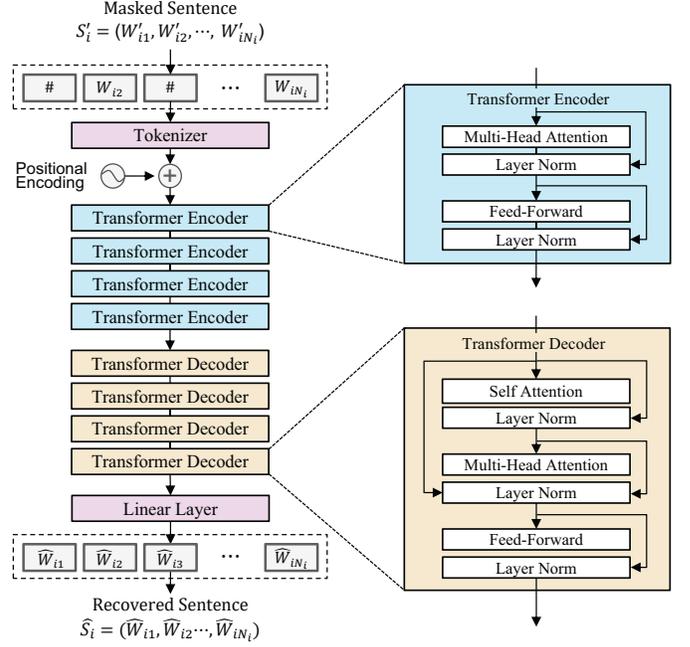}
\caption{Flowchart illustration of the semantic decoder.}
\label{fig:RN}
\end{figure}

\subsection{Semantic Decoding}

We now consider recovering the text from the compressed bit string $\bx$. First, we convert $\bx$ to the word string $T'$ by means of LZ decoding, which comprises a sequence of raw sentences $(S'_1,S'_2,\ldots,S'_M)$. Notice that each $S'_i$ is the counterpart of the masked $S_i$ at the decoder side. We devise a Transformer-based \cite{vaswani2017attention} demasking module as displayed in Fig.~\ref{fig:RN} in order to recover those masked words in each $S'_i$. 

Specifically, each raw sentence $S'_i$ is decomposed into words (some of which have been masked) and is fed to the tokenizer to yield a \emph{one-hot embedding vector} with the dimension of dictionary size. Zero padding is used if the sentence has fewer than 30 words.
Subsequently, the one-hot embedding vector is transformed into \emph{word embedding} with the dimension of $d=128$, and then the 
the following positional encoding (PE) sequence is added to \emph{word embedding} entry by entry:
\begin{equation}
\texttt{PE}_{\mathrm{pos},z} = \begin{cases}
      \sin\left(10^{-\frac{4z}{d}}\mathrm{pos}\right) & \text{if $z\equiv0\mod 2$}\\
      \cos\left(10^{-\frac{4(z-1)}{d}}\mathrm{pos}\right) & \text{if $z\equiv1\mod 2$}.
    \end{cases},
\end{equation}
for $z=1,2,\ldots,d$, where $\mathrm{pos}$ is the position index of each word within the current sentence $S'_i$. Each raw sentence $S'_i$, with its words all cast to the PE-added embedding vectors, is now fed to the Transformer encoder.  The multi-head attention mechanism of the Transformer encoder is desirable in our case in that it captures the interaction between the target word and its neighboring words. As a result, the features extracted from each word can be recognized in three respects: the positional encoding, the pre-textual word embedding, and the post-textual word embedding, all of which enable the subsequent Transformer decoder to guess the masked words in $S'_i$. The above Transformer network is tuned based on the training data set in an end-to-end fashion to minimize the following cross-entropy:
\begin{multline}
\texttt{CE}(S_{i},\widehat{S}_i) = -\sum_{(n,k)} q_n(V_k) \cdot \log_2p_n(V_k)\\
- \sum_{(n,k)}(1-q_n(V_k)) \cdot \log_2(1-p_n(V_k)),
\end{multline}
where $q_n(V_k)\in\{0,1\}$ is the ground-truth label such that $q_n(V_k)=1$ if $W_{in}=V_k$ and $q_n(V_k)=0$ otherwise, while $p_n(W_{in})\in[0,1]$ is the soft decision that reflects the likelihood of the $n$th word of being $V_k$. After the Transformer has been trained, we decode the bit string as in Algorithm \ref{alg:DEC}.

\begin{algorithm}[t]
\caption{Semantic Data Compression (Decoder Part)}
\label{alg:DEC}
\begin{algorithmic}[1]
\Statex \textbf{Input:} Bit string  $\bx$ 
\State Convert $\bx$ to the masked raw text $T'$ by LZ decoding
\For{each raw sentence $S'_{i}$ in $T'$} 
    \State Convert $S'_{i}$ to multiple one-hot embedding vectors
    \State Incorporate positional encoding 
    \State Extract context features by Transformer encoder
    \State Recover the masked words by Transformer decoder
\EndFor
\State Assemble the text $\widehat{T}$ = $(\widehat{S}_1,\widehat{S}_2,\ldots,\widehat{S}_M)$
\Statex \textbf{Output:} Recovered text $\widehat{T}$ 
\end{algorithmic}
\end{algorithm}

\subsection{Entropy vs. Semantic Loss}
\label{subsec:analysis}

\begin{table}[t]
\small

\renewcommand{\arraystretch}{1.2}
\caption{Stationary distribution of the five possible words in the example considered in Section \ref{subsec:analysis}.}
\begin{center}
\begin{tabular}{|c|c|c|c|c|c|}
\hline
Word & $V_1$ & $V_2$ & $V_3$ & $V_4$ & $V_5$ \\
\hline
Stationary Distribution &  0.15 & 0.15 & 0.20 & 0.25 & 0.25\\
\hline
\end{tabular}
\end{center}
\label{tab:huffman}
\end{table}

\begin{figure}[t]
\centering
\begin{minipage}{0.48\textwidth}
\centering
\includegraphics[width=6.5cm]{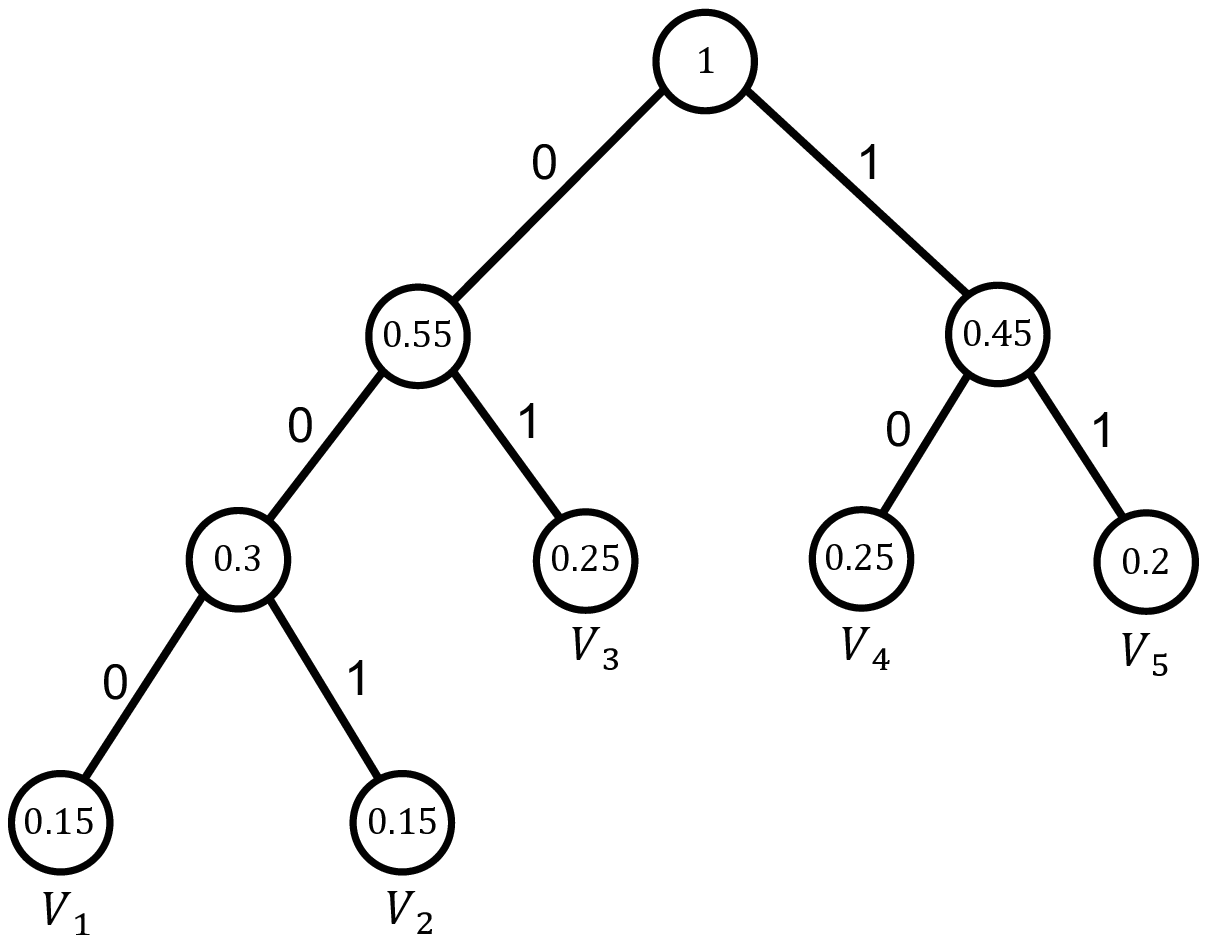}
\caption{Huffman tree for the lossless data compression.}
\vspace{1em}
\label{fig:tree1}
\end{minipage}
\begin{minipage}{0.48\textwidth}
\centering
\hspace{-1em}
\includegraphics[width=5.6cm]{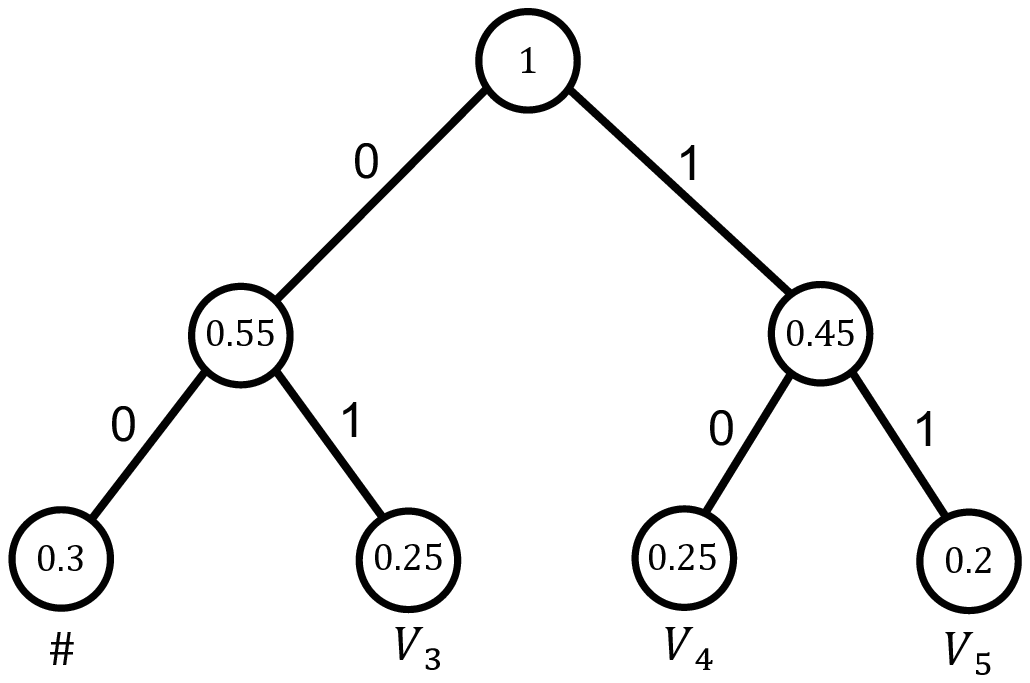}
\caption{Huffman tree after the word merging.}
\label{fig:tree2}
\end{minipage}
\end{figure}

We now present an example to illustrate why the use of semantic loss in \eqref{delta} is critical to the word masking scheme. For ease of discussion, consider a toy model of English text wherein only five possible words $\{V_1,\ldots,V_5\}$ form an \emph{irreducible and aperiodic} five-state Markov chain---which must have the unique stationary distribution. Table \ref{tab:huffman} shows the stationary distribution of the above model. Thus, if we aim at lossless data compression, the optimum (i.e., the minimum expected number of bits for representing the text) can be achieved by Huffman code. One possible construction of the Huffman tree is shown in Fig.~\ref{fig:tree1}.

Now suppose that we wish to further reduce the bit consumption in data compression by choosing two words out of $\{V_1,\ldots,V_5\}$ to merge. Clearly, this would lead us to a lower entropy across the states in the Markov chain because of the data processing inequality. We remark that the word merging can be interpreted as replacing the two chosen words with the special symbol ``\#", namely the word masking. If one pursues the minimum entropy, then it is optimal to merge the two least likely words $V_1$ and $V_2$, as illustrated in Fig.~\ref{fig:tree2}. However, from a semantic point of view, it is often the stop words (e.g., ``a'', ``an'', and ``the'') or the frequently used words in a particular context (e.g., ``learning'' in an academic paper on machine learning) that ought to be masked, so the merged words are expected to have high stationary probabilities. The above discrepancy implies that the conventional entropy metric is not suited for the masking-based data compression.

\section{Experiments}

\begin{figure*}
\small
\fbox{%
\begin{minipage}{56em}
\texttt{\textbf{The} next item \textbf{is} \textbf{the} \textbf{continuation} \textbf{of} \textbf{the} \textbf{joint} debate \textbf{on} agenda. \textbf{Mr} president, commissioners, \textbf{ladies} \textbf{and} \textbf{gentlemen}, \textbf{I} \textbf{would} \textbf{like} \textbf{to} \textbf{focus} \textbf{on} the European social fund. \textbf{I} \textbf{am} voting \textbf{in} \textbf{favour} \textbf{of} \textbf{the} amendments \textbf{proposed} \textbf{by} \textbf{the} Trakatelli’s report \textbf{on} \textbf{the} regulation \textbf{of} leaf tobacco. \textbf{The} next \textbf{is} \textbf{the} report from \textbf{Sir} Jack Stewart Clark \textbf{on} \textbf{behalf} \textbf{of} \textbf{the} committee \textbf{about} civil liberties and internal \textbf{affairs} \textbf{on} \textbf{the} draft \textbf{joint} actions.}
\end{minipage}
}
\caption{A sample paragraph from the 2005 proceedings of the European Parliament \cite{koehn-2005-europarl}. The bold words are to be masked by Algorithm \ref{alg:EN}.}
\label{fig:para1}
\vspace{1em}
%
%
\fbox{%
\begin{minipage}{56em}
 \texttt{\textbf{The} next item \textbf{is the continuation of the joint} debate \textbf{on} agenda. \textbf{Mr} president, commissioners,  \textbf{ladies and gentlemen}, \textbf{I would like to} \textcolor{red}{\textbf{say that}} the European social fund. \textbf{I am} voting \textbf{in}  \textbf{favour of the} amendments \textcolor{red}{\textbf{submitted}} \textbf{by the} Trakatelli’s report \textbf{on the} regulation \textbf{of} leaf tobacco. \textbf{The} next \textbf{is the} report from \textcolor{red}{\textbf{Mr}} Jack Stewart Clark \textcolor{red}{\textbf{farage}} \textbf{on behalf of} committee \textbf{about} civil liberties \textbf{and} internal \textbf{affairs on the} draft \textcolor{red}{\textbf{general}} actions. }
\end{minipage}
}
\caption{Demasking of the above paragraph by Algorithm \ref{alg:DEC}. The recovered words are in bold font; the discrepancies are highlighted in red.}
\label{fig:para3}
\end{figure*}

\subsection{Setup}

\textbf{Data Set:} The sample text used in our experiments is taken from the 2005 proceedings (English version) of the European Parliament \cite{koehn-2005-europarl}. It is composed of around 80,000 sentences with over 1.5 million words in total. The size of each sentence ranges from 4 words to 30 words. The data set is divided into two groups: 90\% for training and the rest 10\% for testing.



\textbf{Hyper-Parameters:} Our Transformer model is trained for 120 epochs until it fully converges. We use the Adam optimizer \cite{kingma2014adam} with the parameter setting $1\times10^{-4}$, $\beta = (0.9,0.98)$, $\epsilon = 1\times10^{-8}$, and a weight decay of $5\times10^{-4}$. 

\textbf{Competitor Algorithms:} Aside from the proposed data compression method, we include in our experiments the traditional lossless compression methods: Huffman coding \cite{huffman1952method} and UTF-8 coding \cite{yergeau2003utf}. We also include a recently proposed semantic data compression method based on deep learning in \cite{niu2022paradigm}. We remark all the methods are parameterized based on the training set, e.g., the codebook of Huffman codes is constructed by evaluating the frequency of each word within the training text data. Moreover, we consider two variations of the proposed algorithm:
\begin{itemize}
    \item \emph{Masking with Long Sentences:} This method combines sentences into long sentences and computes cosine distance in \eqref{delta} between the long sentences. In our case, the original short sentences contain at most 30 words each, while the long sentences contain 227 to 256 words each.
    \item \emph{Frequency-Based Masking:} Rather than using the attention weight $\mu_k$ in \eqref{mu} to sort the words, this method adopts the frequency metric, i.e., those words with the highest frequencies are masked. The motivation behind this new masking scheme is discussed in Section \ref{subsec:analysis}.
\end{itemize}





\subsection{Test Results}

We begin with the proposed semantic compression of a sample paragraph with the masking ratio $\rho=0.674$, as shown in Fig.~\ref{fig:para1} and Fig.~\ref{fig:para3}. Observe that a large portion of masking is applied to the stop words like ``of'', ``is'', ``to'', ``the'' etc. as expected in Section \ref{subsec:analysis}. It is worth remarking that the words and phrases closely related to the context, such as ``affairs'', ``ladies and gentlemen'', and ``proposed'', are masked as well. Observe also from Fig.~\ref{fig:para3} that the recovery is fairly successful since most missing words can be recovered. But there are still some discrepancies, some of which even cause semantic misunderstanding, e.g., ``Sir'' is replaced with ``Mr'', but the former actually indicates an honorific title rather than a regular title before a man's surname. This issue shall be fixed by the background knowledge supplement at the training stage.

\begin{figure}[t]
\centering
\vspace{-2em}
\includegraphics[width=8.6cm]{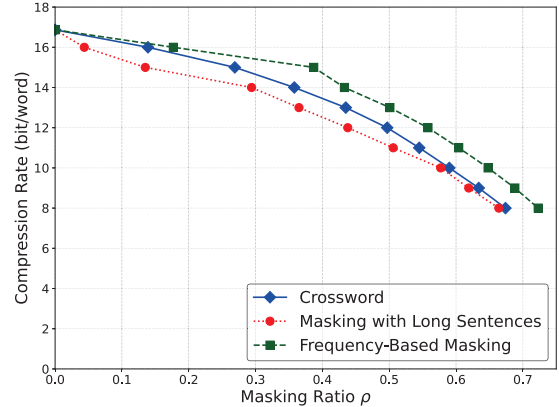}
\caption{Three masking methods under the different masking ratios $\rho$.}
\vspace{-2em}
\hspace{-2.0em}
\label{fig:loss}
\end{figure}

\begin{table*}[t]
\small
\renewcommand{\arraystretch}{1.2}
\caption{The semantic similarity performance of the different lossy compression methods at the same compression rate.}
\begin{center}
\label{tlb:compare}
\begin{tabular}{|c|c|c|c|c|c|c|c|c|c|}
\hline
\diagbox{Methods}{Sentence Similarity}{Bits/word}  & $8.0$ & $9.0$ & $10.0$ & $11.0$ & $12.0$ & $13.0$ & $14.0$ & $15.0$ & $16.0$  \\
\hline
Semantic Compression Method in \cite{niu2022paradigm} & 0.500 & 0.583 & 0.639 & 0.709 & 0.748 & 0.798 & 0.795 & 0.830 & 0.847 \\
\hline
Proposed Method ``Crossword'' & 0.869&  0.893 & 0.913 & 0.936 & 0.953 & 0.968 & 0.980 & 0.990 & 0.994  \\
\hline
Masking with Long Sentences &  0.837 & 0.869 & 0.893 & 0.909 & 0.931 &  0.941 & 0.952 & 0.962 & 0.974  \\
\hline
Frequency-Based Masking &0.684& 0.732& 0.768& 0.849& 0.883& 0.921& 0.958& 0.983& 0.997 \\
\hline
\end{tabular}
\end{center}
\end{table*}

Next, we compare the different masking methods for data compression in Fig.~\ref{fig:loss}. It can be seen that all the three methods have their compression rates drop as the masking ratio $\rho$ is increased. At the same $\rho$, the figure shows that masking with long sentences results in the highest compression efficiency. However, this does not suggest that masking with long sentences outperforms other methods since we have not yet considered their semantic performance.


We now tune the masking ratios for the different masking methods in order to compare their semantic performance at the same compression rate. We also consider the existing semantic compression method in \cite{niu2022paradigm} as a benchmark. For the traditional lossless data  compression schemes, Huffman code requires 24.4 bits/word while UTF-8 code requires 47.2 bits/word. The performance of the lossy compression methods is summarized in Table~\ref{tlb:compare}. Notice that when the compression rate is high, i.e., when we aim at lossless compression, the conventional frequency-based masking leads to the best performance. Nevertheless, this regime is of limited interest for the semantic study since the symbol-wise recovery is the main consideration when $\rho$ is very low. In contrast, when $\rho$ goes up so as to transcend the entropy limit, the proposed masking scheme starts to take the lead. For instance, when 8 bits are allowed for representing each word on average, the proposed algorithm improves upon the frequency-based masking by over 27\% in terms of cosine similarity. Notice also that the Crossword is superior to masking with long sentences. A possible reason is that computing the semantic loss on a per-short-sentence basis can further take sentence phrasing into account, whereas treating many sentences as a whole would lose this crucial information. Furthermore, notice that the Crossword significantly improves the existing semantic compression method in \cite{niu2022paradigm} by over 73\% at the same compression rate of 8 bits/word.

\begin{figure}[t]
\vspace{-2em}
\centering
\includegraphics[width=8.6cm]{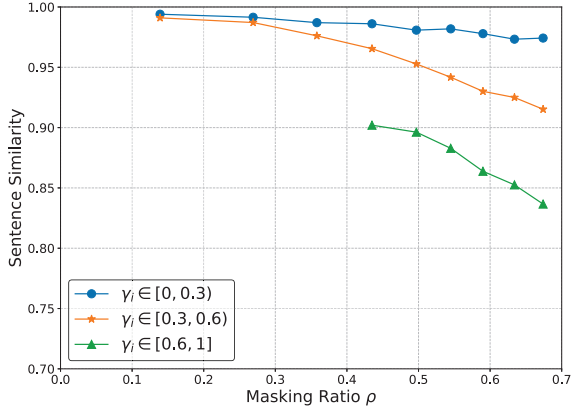}
\caption{Tradeoff between the cosine similarity $\eta(S_i,\hat S_i)$ and the masking ratio $\rho$ for the different groups of sentences.}
\label{fig:f}
\end{figure}

Furthermore, we look into the proposed masking method by characterizing its behavior at the sentence level. Define the \emph{sentence masking ratio} $\gamma_i\in[0,1]$ to be the portion of the words that are masked in the sentence $S_i$. We classify the sentence according to the sentence masking ratio into the following three groups: $\gamma_i \in [0,0.3)$, $\gamma_i \in[0.3,0.6)$, and $\gamma_i \in[0.6,1]$. As shown in Fig.~\ref{fig:f}, when $\rho$ approaches 1 in order to achieve a super high compression efficiency, those sentences with $\rho\in[0.6,1]$ suffer severe semantic loss, whose cosine similarity with the ground truth is only 83\%. But this loss is quickly alleviated when $\rho$ drops. As another important observation, no sentences fall in the group $\gamma_i \in[0.6,1]$ as $\rho$ is below 0.42. The reason is that if $\rho$ gets close to zero then semantic compression gradually reduces to the traditional symbol compression, and thus most sentences end up with their sentence masking ratios $\gamma_i\approx 0$.

\section{Conclusion}

This work proposes a semantic approach to data compression for English text as inspired by the puzzle crossword. The main idea is to mask those semantically minor words in terms of the overall context. The sentence-BERT is used to evaluate the semantic importance while the Transfomer is used to capture the semantic information thereby recovering the masked words. Experiments show the remarkable advantage of the proposed method over the traditional Huffman code and UTF-8 code in minimizing the number of bits for representing text, and also over the existing semantic compression method in \cite{niu2022paradigm} in preserving the meaning of text.


\bibliographystyle{IEEEtran}
\bibliography{IEEEabrv,ref}

\end{document}